\pdfoutput=1

\documentclass[11pt]{article}

\usepackage{subcaption} 
\usepackage[final]{acl}
\usepackage{tabularx}

\usepackage{bbm}
\usepackage{amsmath} 
\usepackage{multirow} 
\usepackage{enumitem}

\usepackage{algorithm}
\usepackage{algpseudocode}
\algtext*{EndIf}
\algtext*{EndFor}
\algtext*{EndWhile}

\usepackage{listings}
\usepackage{amsthm}

\usepackage{xcolor}
\lstdefinestyle{kgprompt}{
  backgroundcolor=\color{gray!10},    
  frame=single,                       
  rulecolor=\color{gray!70},          
  basicstyle=\ttfamily\small,         
  breaklines=true,                    
  columns=fullflexible,
  keepspaces=true,
  showstringspaces=false,
}
\usepackage{times}
\usepackage{latexsym}
\usepackage{booktabs}

\usepackage{url}
\usepackage{todonotes}
\usepackage{amssymb}
\usepackage{amsmath}
\usepackage{algorithm}

\usepackage[T1]{fontenc}

\usepackage[utf8]{inputenc}

\usepackage{microtype}

\usepackage{inconsolata}

\title{What Breaks Knowledge Graph based RAG? Benchmarking and Empirical Insights into Reasoning under Incomplete Knowledge}

\author{%
Dongzhuoran Zhou\textsuperscript{1,2},
Yuqicheng Zhu\textsuperscript{2,3}, 
Xiaxia Wang\textsuperscript{5},
Hongkuan Zhou\textsuperscript{2,3}, \\
\textbf{Yuan He\textsuperscript{4,5}},
\textbf{Jiaoyan Chen\textsuperscript{6}},
\textbf{Steffen Staab\textsuperscript{3,7},}
\textbf{Evgeny Kharlamov\textsuperscript{1,2}} 
\\
\textsuperscript{1}University of Oslo, 
\textsuperscript{2}Bosch Center for AI, 
\textsuperscript{3}University of Stuttgart, 
\textsuperscript{4}Amazon, 
\\
\textsuperscript{5}University of Oxford, 
\textsuperscript{6}The University of Manchester, 
\textsuperscript{7}University of Southampton\\
\texttt{dongzhuoran.zhou@de.bosch.com}\\
}

\begin{document}
\maketitle
\begin{abstract}
Knowledge Graph-based Retrieval-Augmented Generation (KG-RAG) is an increasingly explored approach for combining the reasoning capabilities of large language models with the structured evidence of knowledge graphs. However, current evaluation practices fall short: existing benchmarks often include questions that can be directly answered using existing triples in KG, making it unclear whether models perform reasoning or simply retrieve answers directly. Moreover, inconsistent evaluation metrics and lenient answer matching criteria further obscure meaningful comparisons. In this work, we introduce a general method for constructing benchmarks and present \textbf{BRINK (Benchmark for Reasoning under Incomplete Knowledge)} to systematically assess KG-RAG methods under knowledge incompleteness.
Our empirical results show that current KG-RAG methods have limited reasoning ability under missing knowledge, often rely on internal memorization, and exhibit varying degrees of generalization depending on their design.
\end{abstract}

\section{Introduction}
Retrieval-Augmented Generation (RAG) has become a widely adopted framework for enhancing large language models (LLMs) by incorporating external knowledge through a retrieve-then-generate paradigm \citep{Khandelwal2020rag, Izacard2021FiD, Borgeaud2022retro, ram2023ICRALM, zhu2025argrag}. By conditioning the generation on the retrieved documents, RAG enables LLMs to answer questions or perform tasks using more comprehensive and up-to-date knowledge than what is stored in their parameters.
To improve retrieval accuracy, enable structured reasoning and support explanation, recent research has pivoted toward RAG methods based on Knowledge Graph (KG-RAG) \citep{han2024retrieval,peng2024graph}. Most of them directly use existing knowledge graphs (KGs) \citep{linhao2024rog, Jiashuo2024ToG, zhou2025gr}, while some of them construct and extend structured knowledge from unstructured documents \citep{fang2024reano}.  
Such KG-RAG systems are expected to be well-suited for structured reasoning, where answers require synthesizing information from multiple connected facts.

Despite growing interest, current evaluation practices for KG-RAG fall short in two key ways.
First, most existing benchmarks ~\citep{Yih2016webqsp, Talmor2018WSQ} are constructed on top of complete KGs, where direct evidence supporting the answer is readily available. For example, given the question “\emph{Who is the brother of Justin Bieber?}”, the KG contains the triple \texttt{hasBrother(JustinBieber, JaxonBieber)}, allowing the system to answer the question without reasoning. 
However, real-world KGs are often incomplete, and answering such questions in practice may require reasoning over alternative paths, e.g., combining \texttt{hasParent(JustinBieber, JeremyBieber)} and \texttt{hasChild(JeremyBieber, JaxonBieber)} to infer the sibling relationship. As a result, current benchmarks do not assess whether KG-RAG methods can reason over missing knowledge or simply retrieve answers directly from explicit evidence.

Second, evaluation protocols across KG-RAG studies lack standardization and rigor. We identify two pervasive issues: (1) ambiguous definitions, where metrics like ``accuracy'' fluctuate arbitrarily between Exact Match and permissive substring inclusion; and (2) implementation discrepancies, where official codebases frequently contradict their paper descriptions. For instance, several benchmarks describe using ranking-based metrics (e.g., Hits@1) but implement them as loose existence checks without any ranking. These methodological lapses result in inflated performance estimates and render cross-paper comparisons unreliable.

In this work, we introduce \textbf{BRINK (Benchmark for Reasoning under Incomplete Knowledge)}. Constructed using a novel generalizable method, BRINK includes an evaluation protocol designed to systematically assess KG-RAG methods under knowledge incompleteness.
Each question in our benchmark is constructed such that it \emph{cannot} be answered using a single explicit triple. Instead, the answer must be inferred by reasoning over alternative paths in the KG.
To construct this benchmark, we follow a two-step process:
(1) we mine high-confidence logical rules from the KG using a rule mining algorithm, and
(2) we generate natural language questions based on rule groundings, ensuring that the directly supporting triple is removed while the remaining KG still contains sufficient evidence to infer the answer.

Our empirical study on BRINK reveals several key limitations of current KG-RAG systems. 
First, most models struggle to recover answers when direct supporting facts are removed, highlighting their limited reasoning capacity. Second, training-based methods (e.g., RoG, GNN-RAG) show stronger robustness under KG incompleteness compared to non-trained systems. Third, we find that textual entity labels substantially boost performance, suggesting that LLMs rely heavily on internal memorization.

\section{Related Work}
\subsection{Knowledge Graph Question Answering (KGQA)}
The KGQA task aims to answer natural language questions using the KG~$\mathcal{G}$, where $\mathcal{G}$ is represented as a set of binary facts $r(s, o)$, with $r$ denoting a predicate and $s, o$ denoting entities. The answer to each question is one or more entities in $\mathcal{G}$.

Approaches to KGQA can be broadly categorized into the following two types: (1) \textit{Semantic parsing-based methods}~\cite{Yih2016webqsp,gu2021beyond,ye2021rng} translate questions into formal executable queries (e.g., SPARQL or logical forms), which are then executed over the KG to retrieve the answer entities. 
These methods offer high precision and interpretability, as the reasoning process is explicitly encoded. However, they face several challenges, including difficulty in understanding semantically and syntactically complex questions, handling diverse and compositional logical forms, and managing the large search space involved in parsing multi-relation queries\cite{lan2021survey}.
(2) \textit{Embedding-based methods}~\cite{yao2019kg,baek2023direct}, by contrast, encode questions and entities into a shared vector space and rank answer candidates based on embedding similarity. While these methods are end-to-end trainable and do not require annotated logical forms, they often struggle with multi-hop reasoning \cite{qiu2020stepwise}, lack interpretability \cite{biswas2023knowledge}, and exhibit high uncertainty in their predictions \cite{zhu-etal-2024-predictive, zhu2025conformalized, zhu-etal-2025-predicate, zhu2025certainty}.

\subsection{KGQA with RAG}
Recently, a new line of work has emerged that integrates large language models more tightly into the KG reasoning process.
These KG-RAG methods go beyond previous KGQA approaches by coupling structured retrieval with generative reasoning capabilities of LLMs. 
RoG~\cite{linhao2024rog} adopts a planning-retrieval-reasoning pipeline, where an LLM generates relation paths as plans, retrieves corresponding paths from the KG, and reasons over them for answer generation. 
G-Retriever~\cite{he2024g} retrieves subgraphs via Prize-Collecting Steiner Tree optimization and provides them to LLMs as soft prompts or text prompts for question answering.
GNN-RAG~\cite{mavromatis2024gnn} uses a GNN to select candidate answers and retrieves shortest paths to them, which are verbalized for LLM-based answer generation.
PoG~\cite{chen2024pog} instead decomposes questions into subgoals and iteratively explores and self-corrects reasoning paths using a guidance-memory-reflection mechanism. 
ToG~\cite{Jiashuo2024ToG} treats the LLM as an agent that interactively explores and aggregates evidence on KGs through iterative beam search, 
StructGPT~\cite{jiang2023structgpt} introduces an interface-based framework that enables LLMs to iteratively extract relevant evidence from structured data such as KGs and reason over the retrieved information to answer complex questions.

\subsection{KGQA Benchmarks}
Existing benchmarks such as WebQSP~\cite{Yih2016webqsp}, CWQ~\cite{talmor2018web}, and GrailQA~\cite{gu2021beyond} are widely used to evaluate KGQA. 
However, during dataset construction, only question-answer pairs that yield a gold SPARQL answer on the reference KG are retained, while any unanswerable questions are discarded~\cite{Yih2016webqsp,talmor2018web,gu2021beyond}. This implicitly assumes that each question can be answered directly using an existing fact in the KG, overlooking the reality that KGs are often incomplete.
To address this, recent works~\cite{xu2024generate,zhou2025evaluating} simulate KG incompleteness by either randomly deleting triples from the whole KG or removing those along the shortest path(s) between the question and answer entities. However, these approaches have a key limitation: they cannot ensure that sufficient knowledge remains in the KG to support answering each question.
This can lead to misleading evaluations, as performance drops may stem from unanswerable questions rather than limitations in the model’s reasoning ability.

\subsection{LLM Reasoning Evaluation}
Evaluation of LLM reasoning typically covers both logical question answering and logical consistency, with benchmarks spanning deduction, multiple-choice, proof generation, and consistency constraints such as entailment, negation, and transitivity. Datasets like LogicBench~\cite{parmar2024logicbench}, ProofWriter~\cite{tafjord2020proofwriter}, and LogicNLI~\cite{tian2021diagnosing} are widely used for these tasks, but results indicate that LLMs still struggle with robust logical reasoning~\cite{cheng2025empowering, he2025supposedly}.

\section{Benchmark Construction}
\label{sec:da_construct}
This section presents a general method for constructing benchmarks, evaluating KG-RAG methods that can support different settings of knowledge incompleteness.
The key objective is to create natural language questions whose answers are not directly stated in the KG but can be logically inferred through reasoning over alternative paths.

To achieve this, we first mine high-confidence logical rules from the KG to identify triples that are inferable via reasoning. We then remove a subset of these triples while preserving the supporting facts required for inference. Natural language questions are generated based on the removed triples, meaning that models must rely on reasoning rather than direct retrieval to answer the questions.

\subsection{Rule Mining}
\label{sec:rule_mining}
To ensure that questions in our benchmark require reasoning rather than direct lookup, we first identify triples that are logically inferable from other facts.
We achieve this by mining high-confidence Horn rules from the original KG using the \textsc{AMIE3} algorithm~\citep{lajus2020fast}.

\textsc{AMIE3} is a widely used rule mining system designed to operate efficiently over large-scale KGs. 
A logical rule discovered by \textsc{AMIE3} has the following form (\emph{Horn rules}~\cite{horn1951sentences}):
\begin{equation*}
B_1 \wedge B_2 \wedge \dots \wedge B_n \Rightarrow H\,,
\label{eq:horn}
\end{equation*}
where each item is called an \emph{atom}, a binary relation of the form $r(X,Y)$, in which $r$ is a predicate and $X,Y$ are variables. The left-hand side of the rule is a conjunction of \emph{body atoms}, denoted as $\mathbf{B}=B_1\wedge\dots\wedge B_n$, and the right-hand side is the \emph{head atom} $H$. Intuitively, a rule expresses that if the body $\mathbf{B}$ holds, then the head $H$ is likely to hold as well.

A \emph{substitution} $\sigma$ maps every variable occurring in an atom to an entity that exists in $\mathcal{G}$.
For example applying $\sigma=\{X\mapsto \texttt{Justin}, Y\mapsto \texttt{Jaxon}\}$ to the atom \texttt{hasSibling(X,Y)} yields the grounded fact \texttt{hasSibling(Justin,Jaxon)}. A \emph{grounding} of a rule $\mathbf{B}\Rightarrow H$ is 
\begin{equation*}
    \sigma(B_1)\wedge\dots\sigma(B_n)\Rightarrow\sigma(H)\,.
\end{equation*}

\paragraph{Quality Measure.} \textsc{AMIE3} uses the following metrics to measure the quality of a rule:
\begin{itemize}
    \item \textbf{Support.} The \emph{support} of a rule is defined as the number its groundings for which all grounded facts are observed in the KG:
    \begin{align*}
        &support(\mathbf{B}\Rightarrow H) = \\
        &|\{\sigma(H)\mid \forall i,\sigma(B_i)\in\mathcal{G}\wedge\sigma(H)\in\mathcal{G}\}|\,.
    \end{align*}
 
    \item \textbf{Head coverage.} \emph{Head coverage (hc)} measures the proportion of observed head groundings in the KG that are successfully explained by the rule. It is defined as the ratio of the rule’s support to the number of head groundings in the KG:
    \begin{equation*}
        hc(\mathbf{B}\Rightarrow H) = \frac{support(\mathbf{B}\Rightarrow H)}{|\{\sigma\mid\sigma(H)\in\mathcal{G}\}|}\,.
    \end{equation*}
    \item \textbf{Confidence.} \emph{Confidence} measures the proportion of body groundings that also lead to the head being observed in the KG. It is defined as the ratio of the rule's support to the number body groundings in the KG:
    \begin{equation*}
        confidence(\mathbf{B}\Rightarrow H) = \frac{support(\mathbf{B}\Rightarrow H)}{|\{\sigma\mid\sigma(\mathbf{B})\in\mathcal{G}\}|}\,.
    \end{equation*}
\end{itemize}

We retain only rules with high confidence and sufficient support, filtering out noisy or spurious patterns. Specifically, we run \textsc{AMIE3} with a confidence threshold of 0.3, a head coverage threshold of 0.1, and a maximum rule length of 4. \textsc{AMIE3} incrementally generates candidate rules via breadth-first refinement~\cite{lajus2020fast} and evaluates them using confidence and head coverage; only those meeting the specified thresholds are retained. Additional details on the rule generation and filtering process are provided in Appendix~\ref{app:rule_mining}.

\subsection{Dataset Generation}
We aim to generate questions that cannot be answered using direct evidence, but for which sufficient information is implicitly available in the KG. The core idea is to first remove triples that can be reliably inferred using high-confidence rules mined by \textsc{AMIE3}, and then generate questions based on these removed triples.

\paragraph{Triple Removal.}
\label{sec:kg-deletion}
For each mined rule $\mathbf{B}\Rightarrow H$, 
we extract up to 30 groundings $\sigma(\mathbf{B}\Rightarrow H)$ such that both the grounded body $\sigma(\mathbf{B})$ and grounded head $\sigma(H)$ exist in the KG. 
For each such grounding, we remove the head triple $\sigma(H)$ from the KG while preserving all body triples $\sigma(\mathbf{B})$. 
To ensure that the removed triples remain logically inferable from the remaining KG, we enforce the following two constraints:
\begin{itemize}
    \item All grounded body triples required to infer a removed head must remain in the KG.
    \item Removing a head triple must not eliminate any body triple used by other selected groundings.
\end{itemize}
This guarantees that every removed triple can be inferred through at least one reasoning path provided by a mined rule.

\paragraph{Question Generation.}
\label{sec:question-prompt}
For each removed triple $r(e_h, e_t)$, we use GPT-4 to generate a natural-language question that asks for the answer entity based on the predicate and a specified topic entity. To promote diversity, we randomly designate either the head $e_h$ or the tail $e_t$ as the topic entity, with the other serving as the answer entity. 
The exact prompt template is provided in Appendix \ref{app:generation_prompt}.

\paragraph{Dataset Balancing.}
KGs typically exhibit a "long-tail" distribution, where a small number of entities participate in a disproportionately large number of triples, while the majority appear only infrequently ~\citep{mohamed2020popularity, chen2023knowledge}.
This imbalance can cause many generated questions to share the same answer entity, leading to biased evaluation.

To reduce answer distribution bias, we apply frequency-based downsampling to the generated questions $\mathcal{Q}$, yielding a more balanced subset $\mathcal{Q}' \subseteq \mathcal{Q}$. As described in Algorithm~\ref{alg:downsample}, for each answer entity $a$, we retain at most $\tau \cdot |\mathcal{Q}|$ questions if $a$ exceeds the frequency threshold $\tau$; otherwise, all associated questions are kept.

\begin{algorithm}
\caption{Downsampling Procedure}
\label{alg:downsample}
\begin{algorithmic}[1]
\Require Question set $\mathcal{Q}$; threshold $\tau \in (0, 1]$
\Ensure Balanced subset $\mathcal{Q}' \subseteq \mathcal{Q}$

\State Let $\mathcal{A} \gets$ set of unique answer entities in $\mathcal{Q}$
\State $\mathcal{Q}' \gets \emptyset$
\ForAll{$a \in \mathcal{A}$}
    \State $\mathcal{Q}_a \gets \{q \in \mathcal{Q} \mid \texttt{answer}(q) = a\}$
    \If{$|\mathcal{Q}_a| > \tau \cdot |\mathcal{Q}|$}
        \State Randomly sample $\mathcal{S}_a \subset \mathcal{Q}_a$ 
        \State of size $\lfloor \tau \cdot |\mathcal{Q}| \rfloor$
    \Else
        \State $\mathcal{S}_a \gets \mathcal{Q}_a$
    \EndIf
    \State $\mathcal{Q}' \gets \mathcal{Q}' \cup \mathcal{S}_a$
\EndFor
\State \Return $\mathcal{Q}'$
\end{algorithmic}
\end{algorithm}

\paragraph{Answer Set Completion.}
Although each question is initially generated based on a single deleted triple, there may exist multiple correct answers in the KG.
For example, the question “\textit{Who is the brother of Justin Bieber?}” may have several valid answers beyond the one used to generate the question (e.g., \texttt{Jaxon Bieber}).
To ensure rigorous and unbiased evaluation, we construct for each question a complete set of correct answers using the full KG before any triple deletions. Specifically, for a given topic entity and predicate, 
we identify all tail entities such that the triple \texttt{(topic, predicate, tail)} exists in the KG.
All such entities are collected as the answer set for that question.

\subsection{BRINK Overview}
\label{sec:data_overview}
\paragraph{Knowledge Graphs.} To support a systematic evaluation of reasoning under knowledge incompleteness, we construct BRINK based on three well established KGs: \textbf{Family} \cite{sadeghian2019drum},  \textbf{FB15k-237} \cite{toutanova2015observed} and \textbf{Wikidata5m} \cite{wang2021kepler}. These datasets differ in size, structure, and domain coverage, enabling evaluation across both synthetic and real-world settings.
\begin{itemize}
    \item \textbf{Family} dataset is a synthetic KG encoding well-defined familial relationships, such as \texttt{father}, \texttt{mother}, \texttt{uncle}, and \texttt{aunt}. It is constructed from multiple family units with logically consistent relation patterns and interpretable schema. 
    \item \textbf{FB15k-237} is a widely used benchmark derived from Freebase \cite{dong2014data}. The graph spans 14,541 entities and 237 predicates, covering real-world domains such as people, locations, and organizations. 
    \item \textbf{Wikidata5m} is a large-scale real-world KG constructed from Wikidata~\citep{vrandevcic2014wikidata}. It contains about 4.6 million entities, 822 relations, and over 20 million triples spanning diverse domains such as people, places, organizations, and events \cite{wang2021kepler}.
\end{itemize}

\paragraph{Mined Rules.}
Table~\ref{tab:rule-stats} summarizes the number of mined rules for each dataset, categorized by rule type.
The listed types (e.g., symmetry, inversion, composition) correspond to common logical patterns, while the \emph{other} category includes more complex or irregular patterns (See Appendix~\ref{app:other_rules} for details). 

\begin{table}[h]
\centering
\resizebox{.49\textwidth}{!}{%
\begin{tabular}{lrrr}
\toprule
\textbf{Rule Type} & \textbf{Family} & \textbf{FB15k-237} & \textbf{Wikidata5m} \\
\midrule
Symmetry: $r(x,y)\Rightarrow r(y,x)$   & 0    & 27 & 19   \\
Inversion: $r_1(x,y)\Rightarrow r_2(y,x)$     & 6   & 50 & 48   \\
Hierarchy: $r_1(x,y)\Rightarrow r_2(x,y)$     & 0    & 76 & 0   \\
Composition: $r_1(x,y)\wedge r_2(y,z)\Rightarrow r_3(x,z)$   & 56   & 343 & 0  \\
Other         & 83 & 570 & 168 \\ 

\midrule
\textbf{Total} & \textbf{145} & \textbf{1,066} & \textbf{235} \\
\bottomrule
\end{tabular}
}
\caption{Statistics of mined rules.}
\label{tab:rule-stats}
\end{table}

\begin{table*}[h]
\centering

\resizebox{\linewidth}{!}{
\begin{tabular}{@{}ll@{}}
\toprule
\textbf{Dataset} & \textbf{Example} \\ 
\midrule

\textbf{Family} & 
\begin{tabular}[t]{@{}l@{}}
\textit{Question:} Who is 139's brother? \quad \textit{Topic Entity:} 139 \\
---\\
\textit{Answer:} [\textcolor{red}{205}, 138, 2973, 2974] \\
\textit{Direct Evidence:} \texttt{brotherOf(139,205)}  \\
\textit{Alternative Paths:} \texttt{fatherOf(139,14) $\wedge$ uncleOf(205,14) $\Rightarrow$ brotherOf(139,205)}  \\
\end{tabular} \\

\midrule

\textbf{FB15k-237} & 
\begin{tabular}[t]{@{}l@{}}
\textit{Question:} What is the currency of the estimated budget for 5297 (Annie Hall)? \quad \textit{Topic Entity:} 5297 (Annie Hall) \\
--- \\
\textit{Answer:} [\textcolor{red}{1109 (United States Dollar)}]  \\
\textit{Direct Evidence:} \texttt{filmEstimatedBudgetCurrency(5297, 1109)}  \\
\begin{tabular}{@{}r@{\;}l@{}}
\textit{Alternative Paths:} & \texttt{filmCountry(5297 (Annie Hall), 2896 (United States of America))} \\
$\wedge$ & \texttt{locationContains(2896 (United States of America), 9397 (New York))} \\
$\wedge$ & \texttt{statisticalRegionGdpNominalCurrency(9397 (New York), 1109 (United States Dollar))} \\
$\Rightarrow$ & \texttt{filmEstimatedBudgetCurrency(5297 (Annie Hall), 1109 (United States Dollar))}
\end{tabular}
\end{tabular} \\

\midrule

\textbf{Wikidata5m} & 
\begin{tabular}[t]{@{}l@{}}
\textit{Question:} What series is 3729461 (Lumia 610) a part of? \quad \textit{Topic Entity:} 3729461 (Lumia 610) \\
--- \\
\textit{Answer:} [\textcolor{red}{1059823 (Nokia Rise)}] \\
\textit{Direct Evidence:} \texttt{partOfSeries(3729461 (Lumia 610), 1059823 (Nokia Rise))}  \\
\begin{tabular}{@{}r@{\;}l@{}}
\textit{Alternative Paths:} & \texttt{follows(2846157 (Nokia Lumia 620), 3729461 (Lumia 610)) $\wedge$ partOfSeries(2846157 (Nokia Lumia 620), 1059823 (Nokia Rise))} \\
$\Rightarrow$ & \texttt{partOfSeries(3729461 (Lumia 610), 1059823 (Nokia Rise))} \\
\end{tabular}
\end{tabular} \\

\bottomrule
\end{tabular}
}
\caption{Examples from BRINK datasets. 
Each instance includes a \emph{natural-language question}, a \emph{topic entity} (provided to the KG-RAG model), and the full \emph{set of correct answers}. The 
\textcolor{red}{red-highlighted} 
answer denotes the \emph{hard answer}—i.e., the one whose supporting triple has been removed in the incomplete KG setting. We also show the corresponding \emph{direct evidence} (the deleted triple) and an \emph{alternative path} derived from a mined rule that enables inference of the answer.
}
\label{tab:qa-examples}
\end{table*}

\paragraph{Datasets in BRINK.}
Each dataset instance consists of (1) a natural-language question, (2) a topic entity referenced in the question, and (3) a complete set of correct answer entities derived from the original KG. 
Table~\ref{tab:qa-examples} presents representative examples from each dataset.
The final question set is randomly partitioned into training, validation, and test sets using an 8:1:1 ratio. This split is applied uniformly across all three datasets to ensure consistency. 

We provide two retrieval sources per dataset:
\begin{itemize}
    \item \textbf{Complete KG}: the original KG containing all triples.
    \item \textbf{Incomplete KG}: a modified version where selected triples, deemed logically inferable via \textsc{AMIE3}-mined rules, are removed (cf. Section~\ref{sec:kg-deletion}).
\end{itemize}

Table~\ref{tab:data-stats} summarizes the number of KG triples and generated questions in each split for all three datasets, under complete and incomplete KG settings. In Appendix~\ref{app:llm_answerability}, we quantitatively verify that removing triples from the incomplete KG does not render the resulting questions unanswerable.

\begin{table}[htbp]
    \centering
    \resizebox{\linewidth}{!}{
    \begin{tabular}{lccccc}
        \toprule
        Dataset & \#Triples & Train & Val & Test & Total Qs \\
        \midrule
        Family-Complete                      & 17,615  & 1,749 & 218 & 198 & 2,165 \\
        Family-Incomplete                    & 15,785  & 1,749 & 218 & 198 & 2,165 \\
        FB15k-237-Complete           & 204,087 & 4,374 & 535 & 540 & 5,449 \\
        FB15k-237-Incomplete         & 198,183 & 4,374 & 535 & 540 & 5,449 \\
        Wikidata5m-Complete           & 20,510,107 & 27,720 & 3,466 & 3,465 & 34,651 \\
        Wikidata5m-Incomplete         & 20,478,006 & 27,720 & 3,466 & 3,465 & 34,651 \\
        \bottomrule
    \end{tabular}}
    \caption{Statistics of BRINK.}
    \label{tab:data-stats}
\end{table}

\section{Evaluation Protocol}
\subsection{Standardization Gaps in Prior Work}
\label{sec:standardization}
A major obstacle in KG-RAG research is the inconsistency of evaluation protocols, which often leads to unfair and misleading comparisons. We identify two primary sources of this inconsistency:

\paragraph{Ambiguity in Metric Definitions.}
Many studies rely on vague textual descriptions such as ``whether the top-1 predicted answer is correct'' for Hits@1 without specifying critical details \citep{linhao2024rog,mavromatis2024gnn, he2024g}. It is often unclear how LLM outputs are parsed, how ``top-1'' is defined in the absence of explicit ranking, and what constitutes correctness (e.g., exact match vs. substring inclusion). Some works omit definitions entirely, citing ``previous work'' without verification \citep{Jiashuo2024ToG, chen2024pog}.

\paragraph{Discrepancies in Implementation.}
We observe significant misalignments between metric descriptions and their codebase implementations. For instance, although \citet{linhao2024rog}, \citet{mavromatis2024gnn}, and \citet{he2024g} describe their primary metric as the ranking-based ``Hits@1'', their official implementations entirely bypass ranking mechanisms. Instead, \citet{linhao2024rog} and \citet{mavromatis2024gnn} adopt a permissive criterion where \emph{any} occurrence of the gold answer within the generated text is counted as correct. Similarly, the definition of ``accuracy'' fluctuates between strict Exact Match and loose substring matching across different studies \citep{Jiashuo2024ToG, chen2024pog, jiang2023structgpt}. Because many works reuse reported numbers under the assumption of comparability, these inconsistencies propagate, rendering cross-paper conclusions unreliable.

To address these issues, we adhere to the fully standardized and formally defined protocol in BRINK.

\subsection{Evaluation Setup}
Given a natural-language question $q\in\mathcal{Q}$, access to a KG, and a topic entity, the model is tasked with returning a set of predicted answer entities $\mathcal{P}_q$. 

Since KG-RAG models typically produce raw text sequences as output, we extract the final prediction set $\mathcal{P}_q$ by applying string partitioning and normalizing, following \citet{linhao2024rog}. 
Details of this postprocessing step are provided in Appendix \ref{app:eval_setting}.
Without specific justification, all entities are represented by randomly assigned indices without textual labels (e.g., “Barack Obama” becomes \texttt{39}) to ensure that models rely solely on knowledge from the KG rather than memorized surface forms.

\subsection{Evaluation Metrics}
\label{sec:eval_mec}
Given a set of test questions $\mathcal{Q}$, 
we denote the predicted answer set and the gold answer set for each question $q\in\mathcal{Q}$ as $\mathcal{P}_q$ and $\mathcal{A}_q$, respectively.
The evaluation metrics are defined as follows:

\noindent\textbf{Hits@\!Any.}  
Hits@\!Any measures the proportion of questions for which the predicted answer set overlaps with the gold answer set, i.e., at least one correct answer is predicted:
\[
\mathrm{Hits@\!Any} = \frac{1}{|\mathcal{Q}|}\sum_{q\in\mathcal{Q}}\mathbbm{1}[\mathcal{P}_q\cap\mathcal{A}_q\not =\varnothing]\,.
\]

\noindent\textbf{Precision} and \textbf{Recall.}  
Precision measures the fraction of predicted answers that are correct, while recall measures the fraction of gold answers that are predicted:
\[
\mathrm{Precision} = \frac{1}{|\mathcal{Q}|}\sum_{q\in\mathcal{Q}}\frac{|\mathcal{P}_q \cap \mathcal{A}_q|}{|\mathcal{P}_q|}\,,
\]

\[
\mathrm{Recall} = \frac{1}{|\mathcal{Q}|}\sum_{q\in\mathcal{Q}}\frac{|\mathcal{P}_q \cap \mathcal{A}_q|}{|\mathcal{A}_q|}\,.
\]

\noindent\textbf{F1-score.}
The F1-score is the harmonic mean of precision and recall, computed per question and averaged across all questions:
\[
\mathrm{F1} = \frac{1}{|\mathcal{Q}|}\sum_{q\in\mathcal{Q}}\frac{2\cdot|\mathcal{P}_q \cap \mathcal{A}_q|}{|\mathcal{P}_q|+|\mathcal{A}_q|}\,.
\]

\noindent\textbf{Hits@Hard.}
We define the \emph{hard answer} for each question, denoted as $a_q\in\mathcal{A}_q$, 
as the specific answer entity selected during the question generation process, i.e., the one whose supporting triple was intentionally removed from the KG.
Hits@Hard measures the proportion of predictions including the hard answer. It is defined as:
\[
\mathrm{Hits@Hard} = \frac{1}{|\mathcal{Q}|}\sum_{q\in\mathcal{Q}}\mathbbm{1}[a_q\in\mathcal{P}_q]\,.
\]

\noindent\textbf{Hard Hits Rate.}
We define the \emph{Hard Hits Rate} (HHR) as the fraction of correctly answered questions (i.e., Hits@Any) that include the hard answer in predictions:
\[
\mathrm{HHR} = \frac{Hits@Hard}{Hits@Any}\,.
\]

\section{Experiments and Results}
\subsection{Overall Performance}
We evaluate six representative KG-RAG methods—RoG~\citep{linhao2024rog}, G-Retriever~\citep{he2024g}, GNN-RAG~\citep{mavromatis2024gnn}, PoG~\citep{chen2024pog}, StructGPT~\citep{jiang2023structgpt}, and ToG~\citep{sun2023think}—on the BRINK datasets introduced in Section~\ref{sec:da_construct}.

\begin{figure*}[t]
    \centering
    \begin{minipage}[t]{0.66\linewidth}
        \centering
        \includegraphics[width=\linewidth]{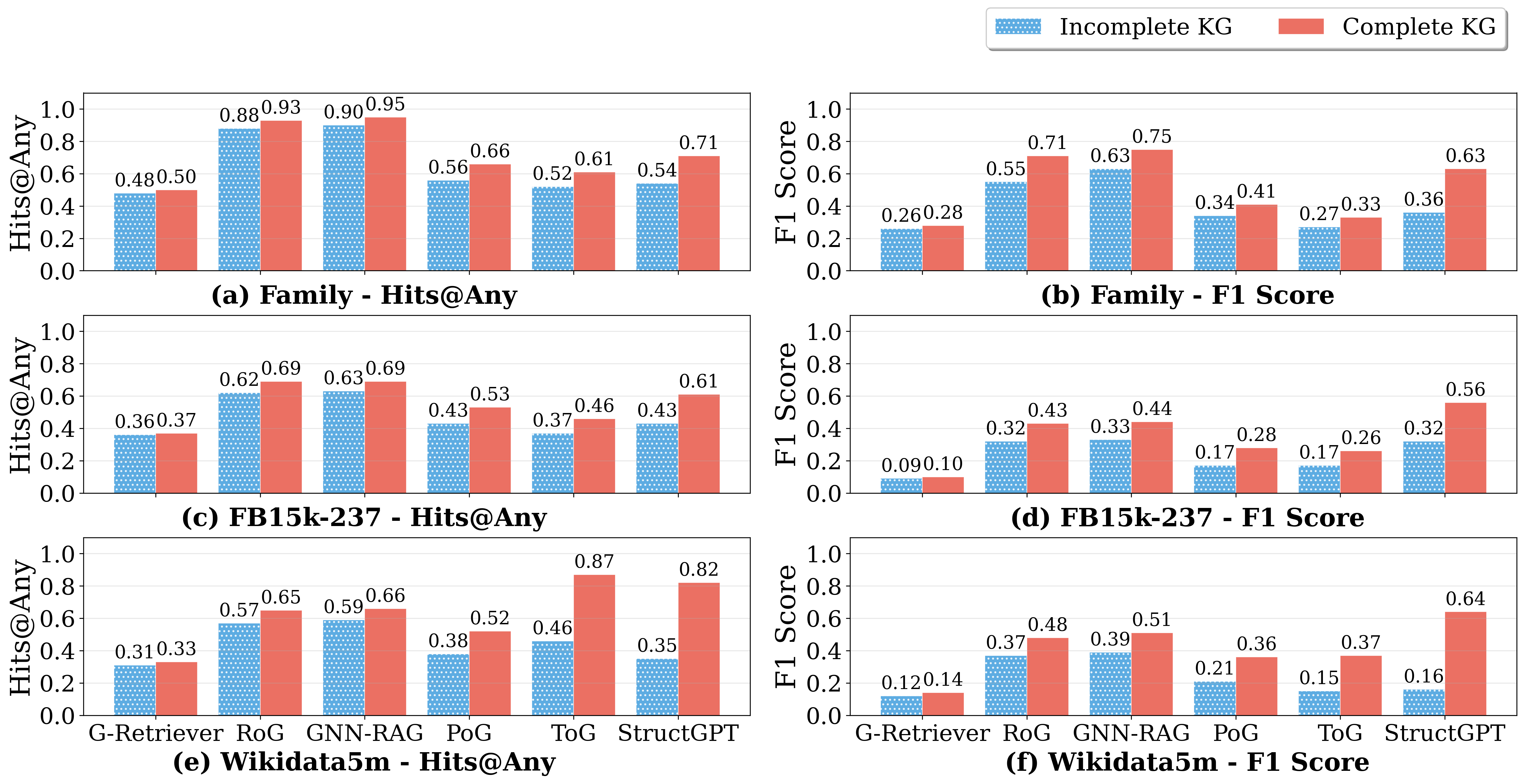}
        \caption{Performance comparison of KG-RAG models under incomplete (blue) and complete (red) KG settings, measured by \textbf{Hits@Any} (left) and \textbf{F1-Score} (right) on KGs: (a-b) Family, (c-d) FB15k-237, (e-f) Wikidata5m.}
        \label{fig:hits_f1}
    \end{minipage}
    \hfill
    \begin{minipage}[t]{0.32\linewidth}
        \centering
        \includegraphics[width=\linewidth]{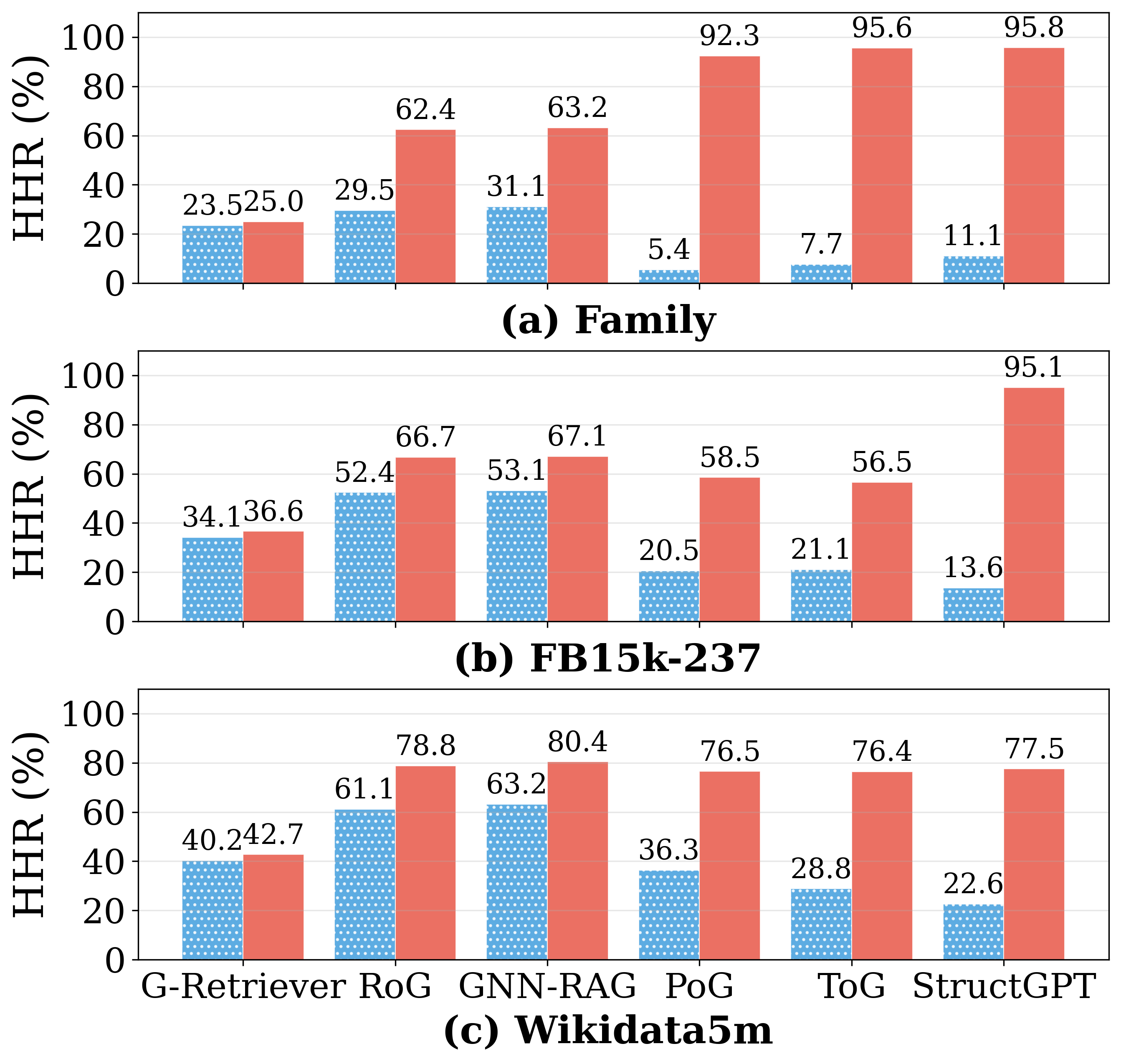}
        \caption{HHR under different KG settings. (a) Family. (b) FB15k-237. (c) Wikidata5m.}
        \label{fig:hard_hits}
    \end{minipage}
\end{figure*}

Figure~\ref{fig:hits_f1} 
reports Hits@Any and F1-scores across all methods and datasets. 
In most cases, \textbf{both metrics drop noticeably when moving from the complete to the incomplete KG setting}, highlighting the challenge posed by missing direct evidence. Precision and recall follow a similar trend (see Appendix~\ref{sec:app_exp}).

G-Retriever presents a unique pattern: 
it shows similarly low performance across both complete and incomplete KGs. This is because its retrieval is based on textual similarity, which often retrieves both topic and answer entities regardless of KG completeness. The GNN encoder can partially recover missing links via neighborhood aggregation, making it less sensitive to missing triples. However, the lack of explicit reasoning and noisy k-NN retrieval lead to irrelevant candidates and overall low F1.

\subsection{Impact of Removing Direct Evidence}
\label{sec:remov_direct}

To examine the impact of removing direct evidence, we report HHR in Figure~\ref{fig:hard_hits}.
Ideally, models capable of reasoning over alternative paths should maintain a similar HHR across both complete and incomplete KG settings.
However, across all models and datasets, \textbf{we observe a significant drop in HHR when moving from the complete to the incomplete KG setting}. This highlights the limited reasoning capabilities of current KG-RAG methods: while they perform well with direct evidence, their effectiveness declines sharply when they need to retrieve alternative paths and reason over it to infer the answer.
Notably, even on the Family dataset, where relation patterns are simple and should be easily recognized by LLMs, models exhibit a substantial decline in recovering the correct answer via alternative paths when direct evidence is absent.

\textbf{Training-based methods (e.g., RoG and GNN-RAG) show a smaller drop in HHR compared to non-trained methods (e.g., PoG and ToG)}, suggesting that exposure to incomplete KGs during training helps models generalize over indirect reasoning paths. In contrast, non-trained methods perform well when direct evidence is available but struggle significantly when such evidence is missing, revealing a stronger sensitivity to incompleteness and more limited reasoning capabilities.

\subsection{Fine-Grained Analysis by Rule Type}
\begin{figure}[t]
    \centering
    \includegraphics[width=0.8\linewidth]{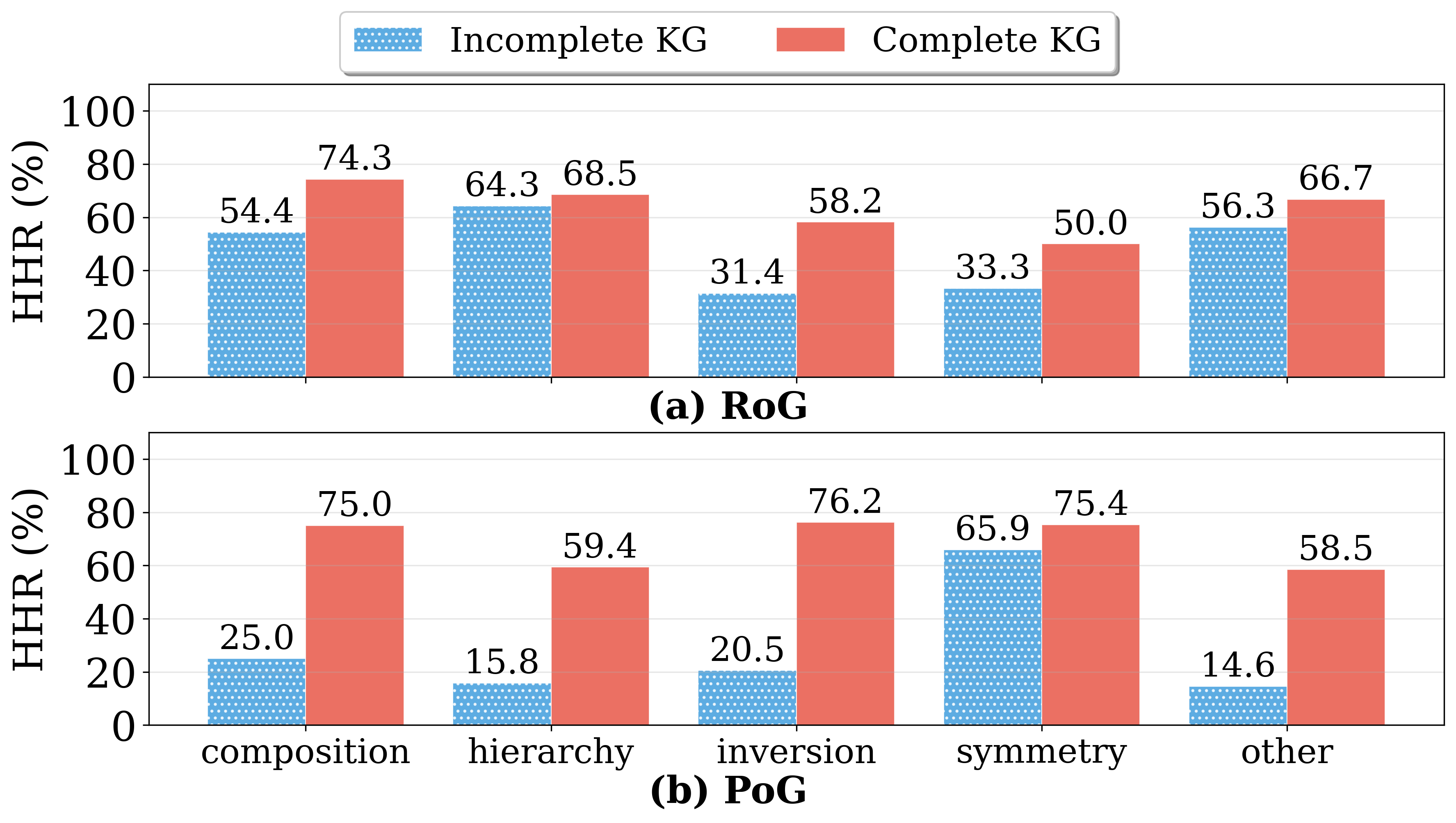}
    \caption{HHR across rule types on FB15k-237 for (a) RoG and (b) PoG, comparing performance under complete and incomplete KG settings.}
    \label{fig:hard_hits_rog_pog}
\end{figure}

To enable a deeper understanding of model reasoning, we conduct a fine-grained analysis of HHR across different rule types on FB15k-237, focusing on two representative models: RoG and PoG (Figure~\ref{fig:hard_hits_rog_pog}).
Overall, RoG exhibits greater robustness to KG incompleteness than PoG across most rule types, indicating that training-based methods can better generalize to multi-hop reasoning when direct evidence is removed. An exception arises in symmetric patterns, where PoG outperforms RoG under the incomplete setting. This is notable because symmetric relations (e.g., \texttt{sibling}) are inherently more robust to directionality. If a model retrieves \texttt{sibling(Bieber, Jaxon)}, it is straightforward for an LLM to infer \texttt{sibling(Jaxon, Bieber)}. RoG’s lower robustness in this seemingly trivial case suggests that training-based methods may overfit to relational patterns seen during training, at the cost of generalizing over simpler relations.

\subsection{Influence of Entity Labeling}

\begin{figure}[t]
    \centering
    \includegraphics[width=1\linewidth]{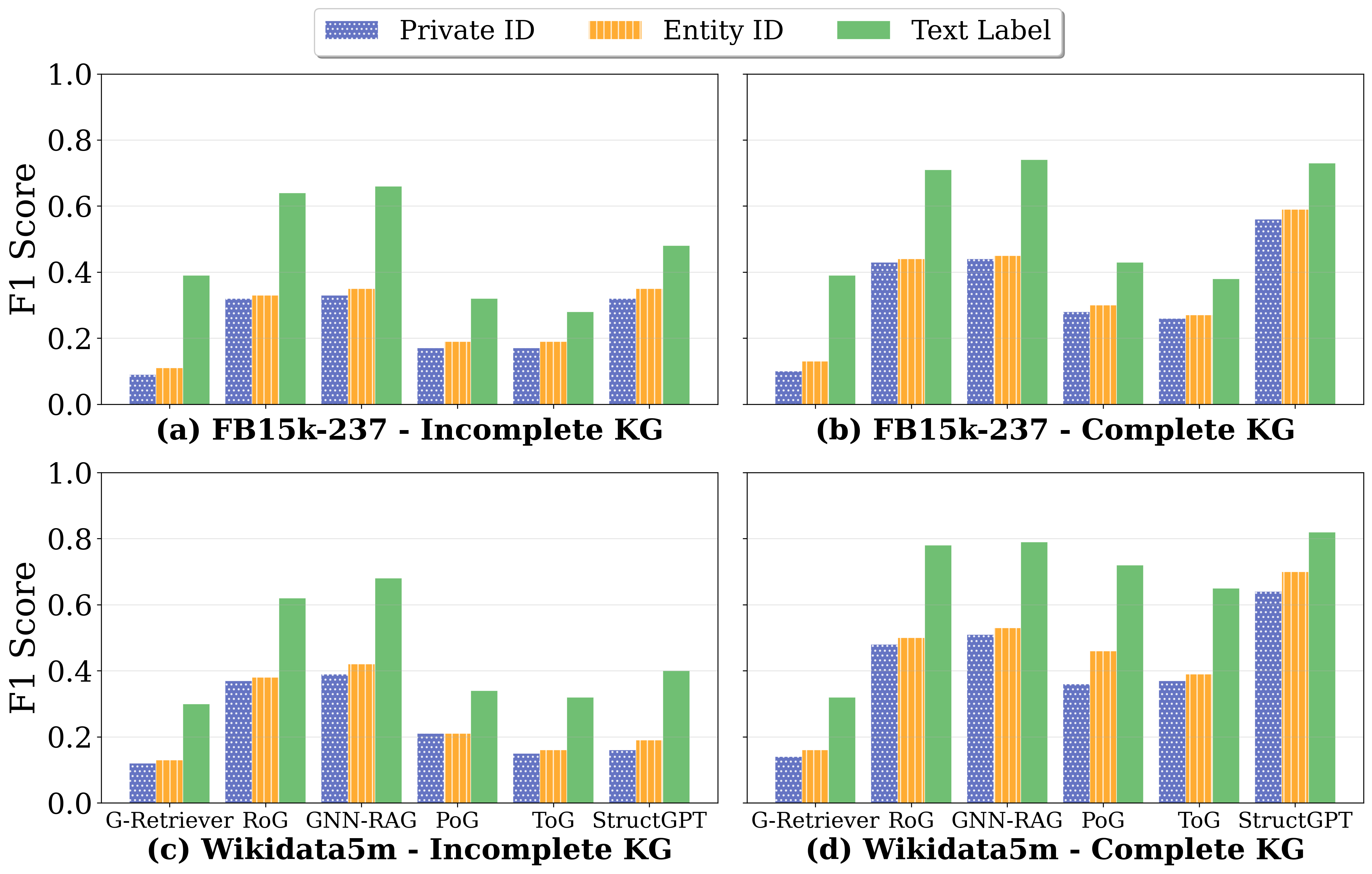}
        \caption{F1-Score comparison of models using Private ID, Entity ID, and Text Label representations on FB15k-237 (a-b) and Wikidata5m (c-d) datasets  under incomplete and complete KG settings.}
    \label{fig:entity_labeling}
\end{figure}

To assess how different entity labeling schemes affect KG-RAG performance, we evaluate three settings for representing entities in the input: (1) \textbf{Private ID}—randomly assigned IDs with no semantic content, (2) \textbf{Entity ID}—official Freebase IDs (e.g., \texttt{/m/02mjmr} for \texttt{Barack Obama}), and (3) \textbf{Text Label}—natural language labels of entities.
Figure~\ref{fig:entity_labeling} shows the F1-scores of all models under each representation, for both incomplete (left) and complete (right) KG settings. We observe two key trends:
First, \textbf{text labels significantly boost performance.} Models consistently achieve higher scores when entity labels are expressed in natural language. This suggests that LLMs can effectively leverage their internal knowledge when text labels are provided. 
Second, \textbf{entity IDs provide limited benefit over random IDs.} Surprisingly, using official entity IDs like \texttt{/m/02mjmr} results in performance nearly identical to that of randomly assigned private IDs. This indicates that, despite LLMs potentially memorizing mappings between surface forms and IDs, they are unable to reason with these identifiers in generation tasks. Instead, they appear to treat IDs as opaque tokens unless the text label is explicit.

\begin{table*}[h]
\centering
\resizebox{\linewidth}{!}{
\begin{tabular}{@{}l@{}}
\toprule

\textbf{Example 1:} \textit{Question:} What is the source of the estimated number of mortgages for \textbf{New York City}? \quad \textit{Answer:} [United States Department of Housing and Urban Development] \\
\textit{Alternative Path:} \texttt{\textcolor{green!60!black}{contains(New York City, Manhattan)} $\wedge$ \textcolor{green!60!black}{source(Manhattan, HUD)}} 
$\Rightarrow$ \texttt{source(New York City, HUD)} \\
--- \\
\textit{Prediction:} [\textcolor{red}{Pace University}] \\
\textit{Retrieved Paths:} 
\quad \texttt{organizationExtra(New York City, Phone Number)}  $\wedge$ \texttt{serviceLocation(Phone Number, \textcolor{red}{Pace University})} \\
\quad \textit{(no path involving mortgage source)} \\
\midrule

\textbf{Example 2:} \textit{Question:} Who is \textbf{Ian Holm}'s spouse? \quad \textit{Answer:} [Penelope Wilton] \\
\textit{Alternative Paths:} \texttt{\textcolor{green!60!black}{spouse(Ian Holm, Penelope Wilton)} $\Rightarrow$ spouse(Penelope Wilton, Ian Holm)} \\
--- \\
\textit{Prediction:} [\textcolor{red}{Marriage}] \\
\textit{Retrieved Paths:} 
\quad \texttt{\textcolor{green!60!black}{spouse(Ian Holm, Penelope Wilton)}} \\
\quad \texttt{awardNominee(Ian Holm, Cate Blanchett) $\wedge$ typeOfUnion(Cate Blanchett, \textcolor{red}{Marriage})} \\
\quad \texttt{awardNominee(Ian Holm, Kate Beckinsale) $\wedge$ typeOfUnion(Kate Beckinsale, Domestic Partnership)} \\
\quad \textit{... (many additional unrelated paths via award nominees)} \\
\midrule
\bottomrule
\end{tabular}
}
\caption{Case study examples from our benchmark illustrating typical failure modes of KG-RAG models. In both examples, the bold text highlights the topic entity, the \textcolor{green!60!black}{green} text denotes the expected alternative path, and the \textcolor{red}{red} text marks incorrect predictions.}
\label{tab:case-study}
\end{table*}

\subsection{Case Study}
\label{sec:case_study}
To better understand the limitations of current KG-RAG methods, we analyze representative failure cases from BRINK. Table~\ref{tab:case-study} shows two illustrative examples: (1) failure to retrieve relevant reasoning paths and (2) incorrect answer generation despite correct retrieval. A quantitative analysis is provided in Appendix~\ref{app:error_analysis}.

\paragraph{(1) Example 1: Retrieval Failure.}
The question requires reasoning over the relations \texttt{contains(New York City, Manhattan)} and \texttt{source(Manhattan, HUD)} to infer the answer \texttt{source(New York City, HUD)}. However, the retriever fails to locate any of these relevant facts. Instead, it returns spurious paths involving entities like \texttt{Pace University}, based on weak co-occurrence signals. As a result, the generator hallucinates an incorrect answer. This type of error is especially common among non-trained retrieval methods, which are more sensitive to missing links and lack robust multi-hop retrieval capabilities.

\paragraph{(2) Example 2: Reasoning Failure.}
The retriever correctly retrieves \texttt{spouse(Ian Holm, Penelope Wilton)}, which should enable inference of the reverse relation. However, the model outputs \texttt{Marriage}, likely influenced by unrelated \texttt{typeOfUnion} paths, indicating difficulty in filtering irrelevant context even with accurate retrieval.

\section{Discussion and Conclusion}
This work introduces a general methodology for constructing benchmarks aimed at evaluating KG-RAG systems under conditions of knowledge incompleteness—a realistic yet often overlooked challenge in real-world KGs. 

Our empirical study on datasets derived from Family, FB15k-237 and Wikidata5m reveals several key limitations of current KG-RAG methods. 
Most notably, existing KG-RAG models struggle to recover answers when direct evidence is missing, indicating limited robustness to incomplete knowledge. While training-based methods exhibit greater resilience, our fine-grained analysis reveals potential overfitting to specific relation patterns, sometimes at the expense of generalizing over trivial structures. 
Furthermore, we find that textual entity labels substantially improve performance, suggesting that models rely more on retrieving memorized knowledge than performing symbolic reasoning over structured data. 


These findings suggest several research directions: (1) developing retrieval strategies that identify alternative paths when evidence is missing, (2) designing reasoning modules with stronger generalization to avoid overfitting to specific relation patterns, and (3) fine-tuning methods that enhance retrieval without impairing reasoning.
\section{Limitation}
Our benchmark construction relies on Horn-style logical rules mined by \textsc{AMIE3} to construct reasoning questions. While this format may appear restrictive, it in fact covers a broad range of reasoning patterns frequently observed in real-world KGs. As shown in Figure \ref{fig:hard_hits_rog_pog}, the benchmark encompasses four major rule types that dominate real-world KGs, with additional diverse patterns summarized in Appendix~\ref{app:other_rules}.

This focus on Horn rules also offers a practical advantage: such rules are interpretable and easy to verify for plausibility, ensuring the reliability and consistency of the constructed questions. In contrast, more expressive rule types are challenging to mine reliably, and even SOTA rule mining methods often produce noisy or implausible rules. To maintain benchmark quality, we therefore focus on Horn rules in this paper. Extending to richer rule types is a promising direction for future work,  specifically, incorporating advanced rule-learning methods like DRUM~\citep{sadeghian2019drum} and AnyBURL~\citep{meilicke2024anytime}, or aggregating high-confidence rules from multiple algorithms, offers a principled way to expand coverage beyond the capabilities of any single system.
\section{Acknowledgements}
The authors thank the International Max Planck Research School for Intelligent Systems (IMPRS-IS) for supporting Yuqicheng Zhu and Hongkuan Zhou. 
The work was partially supported by EU Projects Graph Massivizer (GA 101093202), enRichMyData (GA 101070284), SMARTY (GA 101140087), and the EPSRC project OntoEm (EP/Y017706/1).

\bibliography{custom}

\newpage
\appendix
\section{Details of Rule Mining}
\label{app:rule_mining}

\subsection{AMIE3 Candidate Rule Refinement}
Refinement is carried out using a set of operators that generate new candidate rules:
\begin{itemize}
    \item Dangling atoms, which introduce a new variable connected to an existing one;
    \item Closing atoms, which connect two existing variables;
    \item Instantiated atoms, which introduce a constant and connect it to an existing variable.
\end{itemize}

AMIE3 generates candidate rules by a refinement process using a classical breadth-first search~\cite{lajus2020fast}. It begins with rules that contain only a head atom (e.g. $\Rightarrow \texttt{hasSibling}(X,Y)$) and refines them by adding atoms to the body. For example, it may generate the refined rule:
\begin{align*}
    \texttt{hasParent}(X,Z)&\wedge \texttt{hasChild}(Z,Y)\\
    &\Rightarrow \texttt{hasSibling}(X,Y)\,.
\end{align*}
This refinement step connects existing variables and introduces new ones, gradually building meaningful patterns. 

\subsection{AMIE3 Hyperparameter Settings}
\label{sec:amie3_hyper}

We use AMIE3 with a confidence threshold of 0.3 and a PCA confidence threshold $\theta_{\text{PCA}}$ of 0.4 for all three datasets. The maximum rule length is set to 3 for \textbf{Family} to avoid overly complex patterns, and 4 for \textbf{FB15k-237} and  \textbf{Wikidata5m} to allow richer rules. For the AMIE3 confidence and head coverage thresholds (Section~\ref{sec:rule_mining}), we tested values from 0.1–0.6 and chose 0.3 as a balance: lower values produced unreliable rules, while higher ones biased the distribution toward certain relations. For $\tau$, we tested 0.01–0.2 and set it to limit domination by frequent entities, ensuring a balanced answer distribution.
 See Appendix~\ref{app:pca_conf} for the definition of PCA confidence.

\section{Properties of Horn Rules Mined by AMIE3}
\label{app:amie3_horn_properties}

AMIE3 mines logical rules from knowledge graphs in the form of (Horn) rules:
\[
B_1 \wedge B_2 \wedge \cdots \wedge B_n \implies H
\]
where $B_i$ and $H$ are atoms of the form $r(X, Y)$. To ensure interpretability and practical utility, AMIE3 imposes the following structural properties on all mined rules:

\begin{itemize}
    \item \textbf{Connectedness:} All atoms in the rule are transitively connected via shared variables or entities. This prevents rules with independent, unrelated facts (e.g., $\texttt{diedIn}(x, y) \implies \texttt{wasBornIn}(w, z)$). Two atoms are connected if they share a variable or entity; a rule is connected if every atom is connected transitively to every other atom.
    \item \textbf{Closedness:} Every variable in the rule appears at least twice (i.e., in at least two atoms). This avoids rules that merely predict the existence of some fact without specifying how it relates to the body, such as $\texttt{diedIn}(x, y) \implies \exists z: \texttt{wasBornIn}(x, z)$.
    \item \textbf{Safety:} All variables in the head atom also appear in at least one body atom. This ensures that the rule's predictions are grounded by the body atoms and avoids uninstantiated variables in the conclusion.
\end{itemize}

These restrictions are widely adopted in KG rule mining~\citep{galarraga2015fast,lajus2020fast} to guarantee that discovered rules are logically well-formed and meaningful for downstream reasoning tasks.

\subsection{PCA Confidence}
\label{app:pca_conf}
To understand the concept of rule mining better for the reader we simplified notation of confidence in main body.
Note \textsc{AMIE3} also supports a more optimistic confidence metric known as \emph{PCA confidence}, which adjusts standard confidence to account for incompleteness in the KG. 

\textbf{Motivation.}
Standard confidence for a rule is defined as the proportion of its correct predictions among all possible predictions suggested by the rule. However, this metric is known to be pessimistic for knowledge graphs, which are typically incomplete: many missing triples may be true but unobserved, unfairly penalizing a rule's apparent reliability.

\textbf{Definition.}
To address this, AMIE3 introduces \emph{PCA confidence} (Partial Completeness Assumption confidence)~\citep{galarraga2015fast}, an optimistic variant that partially compensates for KG incompleteness. Given a rule of the form
\[
B_1 \wedge \cdots \wedge B_n \implies r(x, y)
\]
the \textbf{standard confidence} is
\[
\mathrm{conf}(R) = \frac{|\{(x, y): B_1 \wedge \cdots \wedge B_n \wedge r(x, y)\}|}
{|\{(x, y): B_1 \wedge \cdots \wedge B_n\}|}
\]
where the denominator counts all predictions the rule could possibly make, and the numerator counts those that are actually present in the KG.

\textbf{PCA confidence} modifies the denominator to include only those $(x, y)$ pairs for which at least one $r(x, y')$ triple is known for the subject $x$. That is, the rule is only penalized for predictions about entities for which we have observed at least some information about the target relation. Formally,
\begin{align*}
    &\mathrm{conf_{PCA}}(R) = \\
&\frac{|\{(x, y): B_1 \wedge \cdots \wedge B_n \wedge r(x, y)\}|}
{|\{(x, y): B_1 \wedge \cdots \wedge B_n \wedge \exists y': r(x, y')\}|}
\end{align*}

Here, the denominator sums only over those $x$ for which some $y'$ exists such that $r(x, y')$ is observed in the KG.

\textbf{Intuition.}
This approach assumes that, for any entity $x$ for which at least one fact $r(x, y')$ is known, the KG is "locally complete" with respect to $r$ for $x$—so if the rule predicts other $r(x, y)$ facts for $x$, and they are missing, we treat them as truly missing (i.e., as counterexamples to the rule). But for entities where no $r(x, y)$ fact is observed at all, the rule is not penalized for predicting additional facts.

\textbf{Comparison.}
PCA confidence thus provides a more optimistic and fairer assessment of a rule's precision in the presence of incomplete data. It is widely adopted in KG rule mining, and is the default metric for filtering and ranking rules in AMIE3.

For further details, see~\citep{galarraga2015fast}.

\subsection{Rule Mining Procedure}

\begin{algorithm}[h]
\caption{AMIE3}
\label{algo:AMIE}
    \begin{algorithmic}[1]
        \Require Knowledge graph $\mathcal{G}$, maximum rule length $l$, PCA confidence threshold $\theta_{PCA}$, and head coverage threshold $\theta_{hc}$.
        \Ensure Set of mined rules $\mathcal{R}$.
        \State $q \gets$ all rules of the form $\top \Rightarrow r(X, Y)$
        \State $\mathcal{R} \gets \emptyset$
        \While{$q$ is not empty}
            \State $R \gets$ $q$.dequeue()
            \If{$\texttt{SatisfiesRuleCriteria}(R)$}
                \State $\mathcal{R} \gets \mathcal{R} \cup \{R\}$
            \EndIf
            \If{$\texttt{len}(R) < l$ and $\theta_{PCA}(R) < 1.0$}

                \ForAll{$R_c \in \texttt{refine}(R)$}
                    \If{$hc(R_c) \ge \theta_{hc}$ and $R_c \notin q$}
                        \State $q$.enqueue($R_c$)
                    \EndIf
                \EndFor
            \EndIf
        \EndWhile
        \State \Return $\mathcal{R}$
    \end{algorithmic}
\end{algorithm}

\textsc{AMIE3} generates candidate rules by a refinement process using a classical breadth-first search~\cite{lajus2020fast}.
Algorithm~\ref{algo:AMIE} summarizes the rule mining process of \textsc{AMIE3}. 
The algorithm starts with an initial set of rules that contain only a head atom (i.e. $\top \Rightarrow r(X, Y)$, where $\top$ denotes an empty body) and maintains a queue of rule candidates (Line 1). 
At each step, \textsc{AMIE3} dequeues a rule $R$ from the queue and evaluates whether it satisfies three criteria (Line 5):
\begin{itemize}
    \item the rule is \emph{closed} (i.e., all variables in at least two atoms),
    \item its PCA confidence is higher than $\theta_{PCA}$,
    \item its PCA confidence is higher than the confidence of all previously mined rules with the
same head atom as $R$ and a subset of its body atoms.
\end{itemize} 
If these conditions are met, the rule is added to the final output set $\mathcal{R}$. 

If $R$ has fewer than $l$ atoms and its confidence can still be improved (Line 8), \textsc{AMIE3} applies a \texttt{refine} operator (Line 9) that generates new candidate rules by adding a body atom (details in Appendix~\ref{app:rule_mining}). 
Refined rules are added to the queue only if they have sufficient head coverage (Line 11) and have not already been explored.
This process continues until the queue is empty, at which point all high-quality rules satisfying the specified constraints have been discovered.

\subsection{Benchmark Construction Code and Data}

We release the source code for benchmark construction, along with the Family, FB15k-237 and Wikidata5m benchmark datasets, at \url{https://anonymous.4open.science/r/INCK-EA16}.

\section{Additional Results of the experiment}

\subsection{Recall and Precision}
\begin{figure*}[h]
    \centering
    \includegraphics[width=1\linewidth]{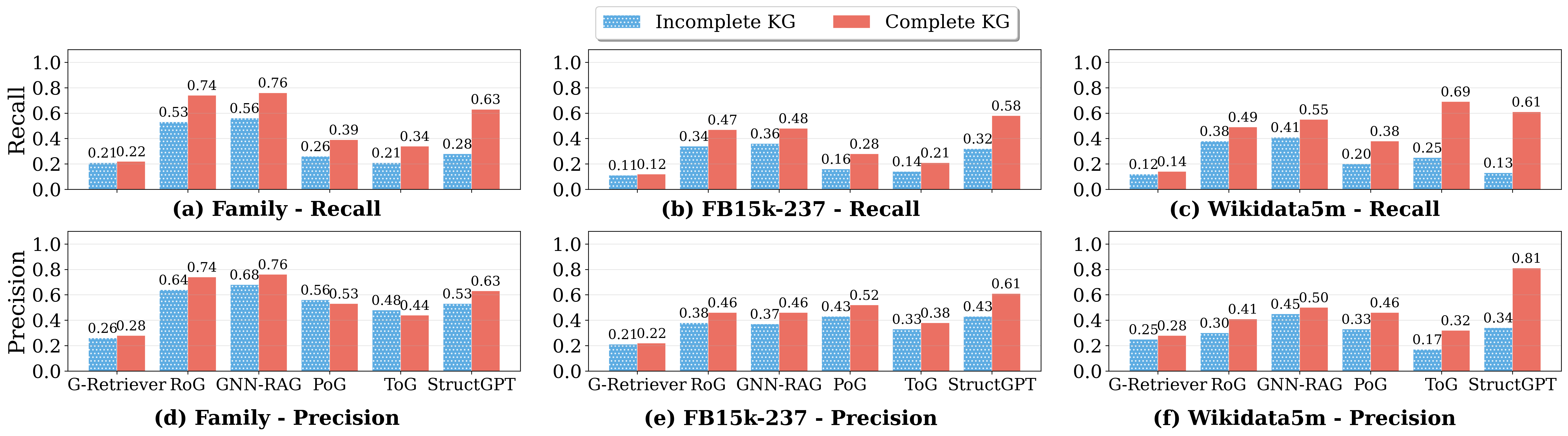}
    \caption{Performance comparison of KG-RAG models under incomplete (blue) and complete (red) KG settings, measured by \textbf{Recall} (top) and \textbf{Precision} (bottom) on Family, FB15k-237, and Wikidata5m.}
    \label{fig:precision_recall}
\end{figure*}

In complement to Figure~\ref{fig:hits_f1}, we provide a detailed breakdown of recall and precision for all settings to offer a more comprehensive evaluation of the F1 scores. Figure~\ref{fig:precision_recall} presents the recall and precision values for all evaluated KG-RAG models across the constructed benchmarks.

\label{sec:app_exp}

\subsection{Evaluation of Generate-on-Graph (GoG)~\citep{xu2024generate}}
This section provides a detailed analysis of GoG~\citep{xu2024generate}, following the experimental results presented in the main text. While GoG explicitly addresses knowledge incompleteness as part of its motivation, we did not include this aspect in the main body for several reasons.

GoG’s implementation is heavily optimized for specific datasets, such as CWQ~\citep{Talmor2018WSQ} and WebQSP~\citep{Yih2016webqsp}, and relies on several domain-specific heuristics:
\begin{itemize}
    \item Hard-coded Templates: Prompting mechanisms and relation-name heuristics tailored to specific KG schemas.
    \item Entity-Type Linking: Reliance on entity-type information that is often unavailable in general-purpose or incomplete KGs.
    \item Textual Dependency: Assumptions regarding entity text labels, which prevents the model from functioning on datasets using anonymized numeric IDs—a necessary measure we take to prevent knowledge leakage and evaluate pure structural reasoning.
\end{itemize}

Despite these limitations, we adapted GoG for our benchmark. The results are shown in Table~\ref{tab:gog_results}. Our evaluation indicates a significant performance decline, particularly in HHR, our metric designed to assess reasoning capacity when direct supporting facts are missing.

The primary limitation lies in GoG’s underlying reasoning mechanism. Rather than performing multi-hop reasoning over alternative paths within the KG, GoG tends to compensate for missing triples by querying the LLM's internal parametric priors. While this strategy may succeed in benchmarks with significant knowledge overlap between the KG and the LLM's training data, it fails in our setting. By anonymizing entity names, we isolate the model’s ability to reason over the provided structure.

\begin{table}[h]
    \centering
    \caption{Performance of GoG on Family and FB15k-237.}
    \label{tab:gog_results}
    \resizebox{\linewidth}{!}{
    \begin{tabular}{lccc}
        \hline
        \textbf{Dataset} & \textbf{Hits@Any} & \textbf{HHR} & \textbf{F1} \\
        \hline
        Family Complete & 0.60 & 0.96 & 0.48 \\
        Family Incomplete & 0.58 & 0.28 & 0.36 \\
        \hline
        FB15k-237 Complete & 0.46 & 0.57 & 0.26 \\
        FB15k-237 Incomplete & 0.20 & 0.23 & 0.11 \\
        \hline
    \end{tabular}
    }
\end{table}

\section{Prompt Template}\label{app:generation_prompt}
Prompt for generating questions from triples:
\begin{lstlisting}[style=kgprompt]
You are an expert in knowledge graph question generation.

Given:
Removed Triple: ({entity_h}, {predicate_T}, {entity_t})
Question Entity: {topic_entity}
Answer Entity: {answer_entity}

Write a clear, natural-language question that asks for the Answer Entity, using the given predicate and Topic Entity.

Requirements:
- Express the predicate {predicate_T} naturally (paraphrasing allowed, but preserve core meaning; e.g., "wife_of" -> "wife").
- Mention the Topic Entity {topic_entity}.
- The answer should be the Answer Entity {answer_entity}.
- Do not mention the Answer Entity {answer_entity} in the question.
- Do not ask a yes/no question.
- Output only the question as plain text.

Example:
Removed Triple: ("Alice", "wife_of", "Carol")
Question Entity: Carol
Answer Entity: Alice

Output:
Who is Carol's wife?

Now, generate the question for:
Removed Triple: ({entity_h}, {predicate_T}, {entity_t})
Question Entity: {topic_entity}
Answer Entity: {answer_entity}
\end{lstlisting}

To ensure reproducibility and mitigate randomness in LLM outputs~\cite{potyka2024robust}, we set the generation temperature to 0 in all experiments.

\section{Detailed Evaluation Settings}\label{app:eval_setting}
All evaluated models are required to produce their predictions as a \emph{structured list of answers}, but in practice, the model output is often a raw string $P_{\mathrm{str}}$ (e.g., \texttt{"Paris, London"} or \texttt{"Paris  London"}). To obtain a set-valued prediction suitable for evaluation, we first apply a splitting function $\mathrm{split}(P_{\mathrm{str}})$, which splits the raw string into a list of answer strings $P = [p_1, p_2, \ldots, p_n]$ using delimiters such as commas, spaces, or newlines as appropriate.

We then define a normalization function $\mathrm{norm}(\cdot)$, which converts each answer string to lowercase, removes articles (\texttt{a}, \texttt{an}, \texttt{the}), punctuation, and extra whitespace, and eliminates the special token \texttt{<pad>} if present. The final prediction set is then defined as $\mathcal{P} = \{\mathrm{norm}(p) \mid p \in P\}$, i.e., the set of unique normalized predictions. The same normalization is applied to each gold answer in the list $A$ to obtain the set $\mathcal{A}$.

All evaluation metrics are computed based on the resulting sets of normalized predictions $\mathcal{P}$ and gold answers $\mathcal{A}$.

\begin{algorithm}
\caption{Output Processing}
\label{alg:normalize}
\begin{algorithmic}[1]
\Require Model output string $P_{\mathrm{str}}$, gold answer list $A$
\Ensure Normalized prediction set $\mathcal{P}$, normalized gold set $\mathcal{A}$
\State $P \gets \mathrm{split}(P_{\mathrm{str}})$
\State $\mathcal{P} \gets \{\, \mathrm{norm}(p) \mid p \in P \,\}$ 
\State $\mathcal{A} \gets \{\, \mathrm{norm}(a) \mid a \in A \,\}$ 
\State \Return $\mathcal{P}, \mathcal{A}$
\end{algorithmic}
\end{algorithm}

\section{Baseline Details}\label{app:baseline}
Unless otherwise specified, for all methods we use the LLM backbone and hyperparameters as described in the original papers.

RoG, G-Retriever, and GNN-RAG are each trained and evaluated separately on the 8:1:1 training split of each dataset (Family, FB15k-237 and Wikidata5m) using a single NVIDIA H200 GPU, as described in Section~\ref{sec:data_overview}. For RoG, we use LLaMA2-Chat-7B as the LLM backbone, instruction-finetuned on each dataset for 3 epochs separately. The batch size is set to 4, the learning rate to $2 \times 10^{-5}$, and a cosine learning rate scheduler with a warmup ratio of 0.03 is adopted~\citep{linhao2024rog}. For G-Retriever, the GNN backbone is a Graph Transformer (4 layers, 4 attention heads per layer, hidden size 1024) with LLaMA2-7B as the LLM. Retrieval hyperparameters and optimization follow~\citet{he2024g}. For GNN-RAG~\citep{mavromatis2024gnn}, we use the recommended ReaRev backbone and sBERT encoder; the GNN component is trained for 200 epochs with 80 epochs of warmup and a patience of 5 for early stopping. All random seeds are fixed for reproducibility. 
For PoG~\citep{chen2024pog}, StructGPT~\citep{jiang2023structgpt}, and ToG~\citep{Jiashuo2024ToG}, we use GPT-3.5-turbo as the underlying LLM, following the original papers, and adopt the original prompt and generation settings from each method. The temperature of LLMs is set to 0.

\section{Detailed Analysis of Other Rule Types}
\label{app:other_rules}

The \emph{Other} category in Table~\ref{tab:rule-stats} encompasses a broad range of logical rules that do not fall into standard symmetry, inversion, hierarchy, or composition classes. Below we summarize the main patterns observed, provide representative examples, and discuss their impact on model performance.

\paragraph{Longer Compositional Chains.} 
Rules involving three, 
\begin{align*}
    &r_1(x, y) \wedge r_2(y, z) \wedge r_3(z, w) \Rightarrow r_4(x, w)
\end{align*}

\paragraph{Triangle Patterns.}
Rules connecting three entities in a triangle motif, 
\begin{align*}
    &r_1(x, y) \wedge r_2(x, z) \Rightarrow r_3(y, z)
\end{align*}

\paragraph{Intersection Rules.}
Rules where multiple body atoms share the same argument, 
\begin{align*}
    &r_1(x, y) \wedge r_2(x, y) \Rightarrow r_3(x, y)
\end{align*}

\paragraph{Other  Patterns.}
Some rules do not exhibit simple interpretable motifs, involving unusual variable binding or rare predicate combinations. Like recursive rules (check AMIE3~\cite{lajus2020fast} for more details)




\section{Quality Check of the Generated Questions}
\label{app:llm_answerability}

To ensure that the constructed benchmark does not produce unanswerable questions, we conduct a quality check on the generated question set. Although the construction process guarantees that all grounded body triples required to infer a removed head remain in the KG, it is still necessary to verify that the removed triples can indeed be inferred from the remaining facts. Ensuring the answerability of generated questions is therefore essential, and a quantitative verification is required to confirm that the removed triples remain inferable from the retained facts.

To examine this, we evaluate whether each removed triple can be recovered when the model is given sufficient contextual information. Specifically, we provide both the alternative reasoning paths and the corresponding logical rules for each question, thereby simulating a perfect retriever with guided reasoning. This setup ensures that all necessary evidence for deriving the target entity is explicitly available. Performance is reported as the proportion of cases in which the target entity was correctly predicted.

The results in Table~\ref{tab:sufficient_info} show that, under this setting, LLMs answer most questions correctly, indicating that the benchmark questions are indeed answerable by LLMs when sufficient information is available.

\begin{table}[h]
\centering
\caption{Performance of backbone LLMs when provided with sufficient information}
\label{tab:sufficient_info}
\resizebox{\linewidth}{!}{
\begin{tabular}{lccc}
\toprule
\textbf{Backbone} & \textbf{Family} & \textbf{FB15k-237} & \textbf{Wikidata5m} \\
\midrule
GPT-3.5-turbo & 0.88 & 0.86 & 0.84 \\
GPT-4o & 0.91 & 0.94 & 0.90 \\
\bottomrule
\end{tabular}
}
\end{table}

\section{Results under Permissive Evaluation Metrics}
\label{app:old_metrics}

We also report the performance of representative KG-RAG models under the commonly used permissive metrics adopted in prior works, such as Hits@1 and relaxed F1/Recall/Precision~\citep{linhao2024rog,chen2024pog}. These metrics treat partial string matches as correct answers and may therefore overestimate model performance, for example by counting a response that contains the gold entity in a negated form as correct (e.g., treating ``not Barack Obama'' as correct when the gold answer is ``Barack Obama''). Nevertheless, as shown in Table~\ref{tab:old_metrics}, the results remain consistent with those obtained using our stricter evaluation protocol, revealing substantial performance differences between complete and incomplete KGs.

\begin{table}[h]
\centering
\caption{Performance of representative KG-RAG models under standard permissive metrics.}
\label{tab:old_metrics}
\resizebox{\linewidth}{!}{
\begin{tabular}{lccccc}
\hline
\textbf{Method} & \textbf{Settings} & \textbf{Hits@1} & \textbf{Permissive F1} & \textbf{Permissive Recall} & \textbf{Permissive Precision} \\
\hline
RoG & complete KG & 69.54 & 43.97 & 47.25 & 46.08 \\
RoG & incomplete KG & 62.97 & 33.24 & 33.63 & 37.81 \\
PoG & complete KG & 57.59 & 30.21 & 27.49 & 56.99 \\
PoG & incomplete KG & 45.26 & 18.56 & 18.47 & 45.66 \\
\hline
\end{tabular}
}
\end{table}

The consistent performance gap across both permissive and strict evaluation settings further validates the robustness of our findings and highlights the distinct behaviors of KG-RAG models when reasoning over incomplete knowledge graphs. In particular, both representative methods (RoG and PoG) exhibit significant performance drops across permissive and strict evaluation settings, and other models show similarly clear declines under incomplete KGs.

\section{Quantitative Error Analysis}
\label{app:error_analysis}

To complement the qualitative case studies presented in Section~\ref{sec:case_study}, we further quantify the distribution of failure types across datasets and methods. Based on the taxonomy introduced in Section~\ref{sec:case_study}, we aggregate the results into two broad categories: \textit{Retrieval Failure} and \textit{Reasoning Failure}. On average, the ratio between these two categories is approximately 7:3, with the majority of failures attributable to Retrieval Failure rather than Reasoning Failure. The ratio is averaged across datasets and models, following the classification criteria in Section~\ref{sec:case_study}: a case is marked as \textit{Retrieval Failure} if no alternative path supporting the gold answer is retrieved, and as \textit{Reasoning Failure} if such a path is retrieved but the final answer is incorrect.

This quantitative summary reinforces the qualitative findings discussed in Section~\ref{sec:case_study}, suggesting that Retrieval Failure remains the primary bottleneck for KG-RAG models operating under incomplete KGs.

\section{Personal Identification Issue in FB15k-237 and Wikidata5m}
While FB15k-237 and Wikidata5m contain information about individuals, they typically focus on well-known public figures such as celebrities, politicians, and historical figures. Since this information is already widely available online and in various public sources, their inclusion in Freebase and Wikidata doesn't significantly compromise individual privacy compared to datasets containing sensitive personal information.

\section{AI Assistants In Writing}
We use ChatGPT \cite{openai2024chatgpt} to enhance our writing skills, abstaining from its use in research and coding endeavors.
\end{document}